\documentclass[letterpaper, 10 pt, conference]{ieeeconf}
\usepackage{times}
\usepackage{graphicx}
\usepackage{amsmath,amssymb,amsopn,amstext,amsfonts}
\usepackage{cancel}
\usepackage[space]{cite}
\usepackage{pdfsync}
\usepackage{balance}
\usepackage{color}
\usepackage{mathtools}
\usepackage{algpseudocode}
\usepackage{algorithm}
\usepackage{bm}

\usepackage{diagbox}
\usepackage{float}
\usepackage{epstopdf}
\usepackage{pifont}
\usepackage{amsmath}
\usepackage{multirow}
\usepackage{url}
\usepackage{verbatim}
\usepackage{booktabs}
\usepackage{graphicx}
\usepackage{subcaption}
\usepackage{threeparttable}
\usepackage{makecell}
\usepackage{soul}
\usepackage{xcolor}
\usepackage[normalem]{ulem}
\usepackage{etoolbox}
\usepackage{url}
\usepackage{mathrsfs}
\usepackage{hyperref}
\usepackage{stfloats}
\hypersetup{hypertex=true,
	colorlinks=true,
	linkcolor=black,
	anchorcolor=black,
	citecolor=black}

\graphicspath{{./figures/}}
\DeclareGraphicsExtensions{.png,.jpg,.eps,.pdf,.emf}
\IEEEoverridecommandlockouts
\newcommand{\delete}[1]{{\bgroup\markoverwith{\textcolor{red}{\rule[0.5ex]{2pt}{0.4pt}}}\ULon{#1}}}
\newcommand{\deletefig}[1]{{\bgroup\markoverwith{\textcolor{red}{\rule[2.5ex]{2pt}{2.0pt}}}\ULon{#1}}}

\setstcolor{red}

\begin{document}
	\title{\LARGE \bf Skater: A Novel Bi-modal Bi-copter Robot for Adaptive \\ Locomotion in Air and Diverse Terrain}
	\author{Junxiao Lin\textsuperscript{2,3}, Ruibin Zhang\textsuperscript{1,2}, Neng Pan\textsuperscript{1,2}, Chao Xu\textsuperscript{1,2}, and Fei Gao\textsuperscript{1,2} 
	\thanks{\textsuperscript{1}Institute of Cyber-Systems and Control, College of Control Science and Engineering, Zhejiang University, Hangzhou 310027, China.}
	\thanks{\textsuperscript{2}Huzhou Institute, Zhejiang University, Huzhou 313000, China.}
	\thanks{\textsuperscript{3}Polytechnic Institute, Zhejiang University, Hangzhou 310015, China.}
	\thanks{Corresponding Author: Fei Gao}
	\thanks{E-mail:{\tt\small \{jxlin, fgaoaa\}@zju.edu.cn}}}

	\maketitle
	\thispagestyle{empty}
	\pagestyle{empty}
	\begin{abstract}
		\label{sec:abstract}\textbf{}
		In this letter, we present a novel bi-modal bi-copter robot called Skater, which is adaptable to air and various ground surfaces. Skater consists of a bi-copter moving along its longitudinal direction with two passive wheels on both sides. Using a longitudinally arranged bi-copter as the unified actuation system for both aerial and ground modes, this robot not only keeps a concise and lightweight mechanism but also possesses exceptional terrain traversing capability and strong steering capacity. Moreover, leveraging the vectored thrust characteristic of bi-copters, the Skater can actively generate the centripetal force needed for steering, enabling it to achieve stable movement even on slippery surfaces. Furthermore, we model the comprehensive dynamics of the Skater, analyze its differential flatness, and introduce a controller using nonlinear model predictive control for trajectory tracking. The outstanding performance of the system is verified by extensive real-world experiments and benchmark comparisons.
	\end{abstract} 
	
	\IEEEpeerreviewmaketitle 
	
	\section{Introduction}
	\label{sec:Introduction}
	In recent years, aerial-ground robots have shown promising results in both academia \cite{kalantari2013hytaq,kalantari2020drivocopter,cao2023doublebee,sihite2023multi} and industry \cite{xpeng,aeromobil,pla-v}. Consisting of hybrid flying and driving mechanisms, they possess distinct advantages over single-modal aerial or ground robots by cross-domain mobility. However, incorporating both locomotion modes in a single robot prototype brings huge challenges to the design of aerial-ground robots. First, the extra weight induced by driving mechanisms decreases flight endurance. This, in turn, limits our choice of driving mechanisms. Moreover, flying mechanisms capable of lifting the entire robot may greatly enlarge the robot size, thus hindering the robot's application in narrow environments. In summary, an impractical design of aerial-ground robots would result in inferior maneuverability in both locomotion modes compared to those of single-modal robots.

	\begin{figure}[ht]  
		\centering
		\includegraphics[width=1.0\columnwidth]{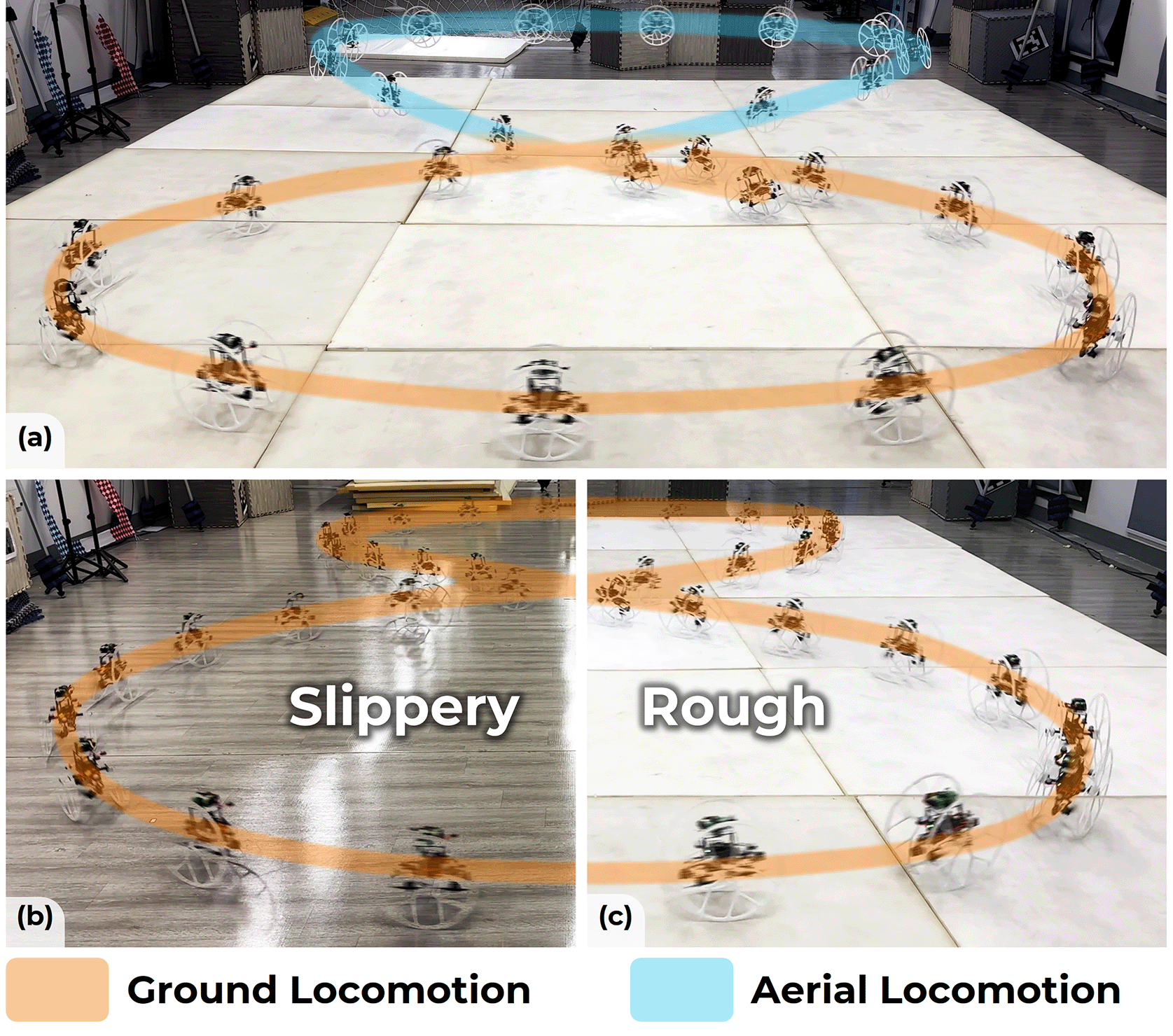}
		\captionsetup{font={small}}
		\caption{ \label{fig:real_world_experiment}Real-world trajectory tracking experiments. (a) Snapshot of aerial-ground hybrid trajectory tracking test. (b-c) Snapshots of trajectory tracking tests on slippery and rough ground surfaces.}
		\vspace{-0.2cm}
	\end{figure}
	
	To address this challenge, researchers have proposed numerous configurations of aerial-ground robots, covering various combinations of flying and driving mechanisms. Firstly, for achieving aerial mobility, multicopters are a common choice for their vertical takeoff and landing capabilities. Among them, quadrotors are widely used due to their compact mechanism and simple dynamics \cite{kalantari2013hytaq,kalantari2020drivocopter,sihite2023multi}. However, the circular arrangement of four rotors in a quadrotor significantly increases its size, and its steering ability is limited by relatively small rotor torques. In contrast, bi-copters, featuring two rotors and servo motors that tilt the rotors, generate turning torque from rotor thrust, which is typically an order of magnitude greater than rotor torque. This enhanced turning torque addresses the issue of limited steering ability in quadrotors. However, horizontally arranged bi-copters still exhibit a relatively wide traversing width \cite{yang2022sytab,cao2023doublebee}. Through comprehensive analysis in section \ref{sec:configuration_design_imple}, we demonstrate that, among common multicopter configurations, bi-copter moving along its longitudinal direction has the smallest traversing width and enhanced steering capability. Therefore, our robot utilizes a longitudinally arranged bi-copter for its flying mechanism.

	Then, regarding ground mobility, wheeled vehicles are widely used as the driving mechanisms for aerial-ground robots due to their concise and lightweight structure \cite{tan2021multimodal,cao2023doublebee,zhang2023model,jia2023quadrolltor,pan2023skywalker}. They can be mainly classified into two types: active-wheeled and passive-wheeled. Active-wheeled robots achieve car-like mobility through the installation of driving wheels on robots \cite{tan2021multimodal,cao2023doublebee}. On the other hand, passive-wheeled robots leverage rotor thrust to drive their passive mechanisms  \cite{zhang2023model,jia2023quadrolltor,pan2023skywalker}. Most active-wheeled robots possess good mobility on rough terrain. However, on slippery surfaces, where the ground fails to provide sufficient friction, these robots are prone to slipping and sliding. To achieve adaptive locomotion on various ground surfaces, we opt for passive-wheeled robots driven by rotor thrust. Most passive-wheeled robots can provide longitudinal acceleration required for forward and backward motion, while they cannot generate the necessary centripetal force for steering, hindering their application to slippery surfaces. Fortunately, longitudinally arranged bi-copter, owing to its vectored thrust characteristic, can generate both longitudinal and lateral accelerations, enabling it to maintain stable and effective movement even on slippery surfaces.

\begin{figure}[t]    
		\centering
		{\includegraphics[width=0.9\columnwidth]{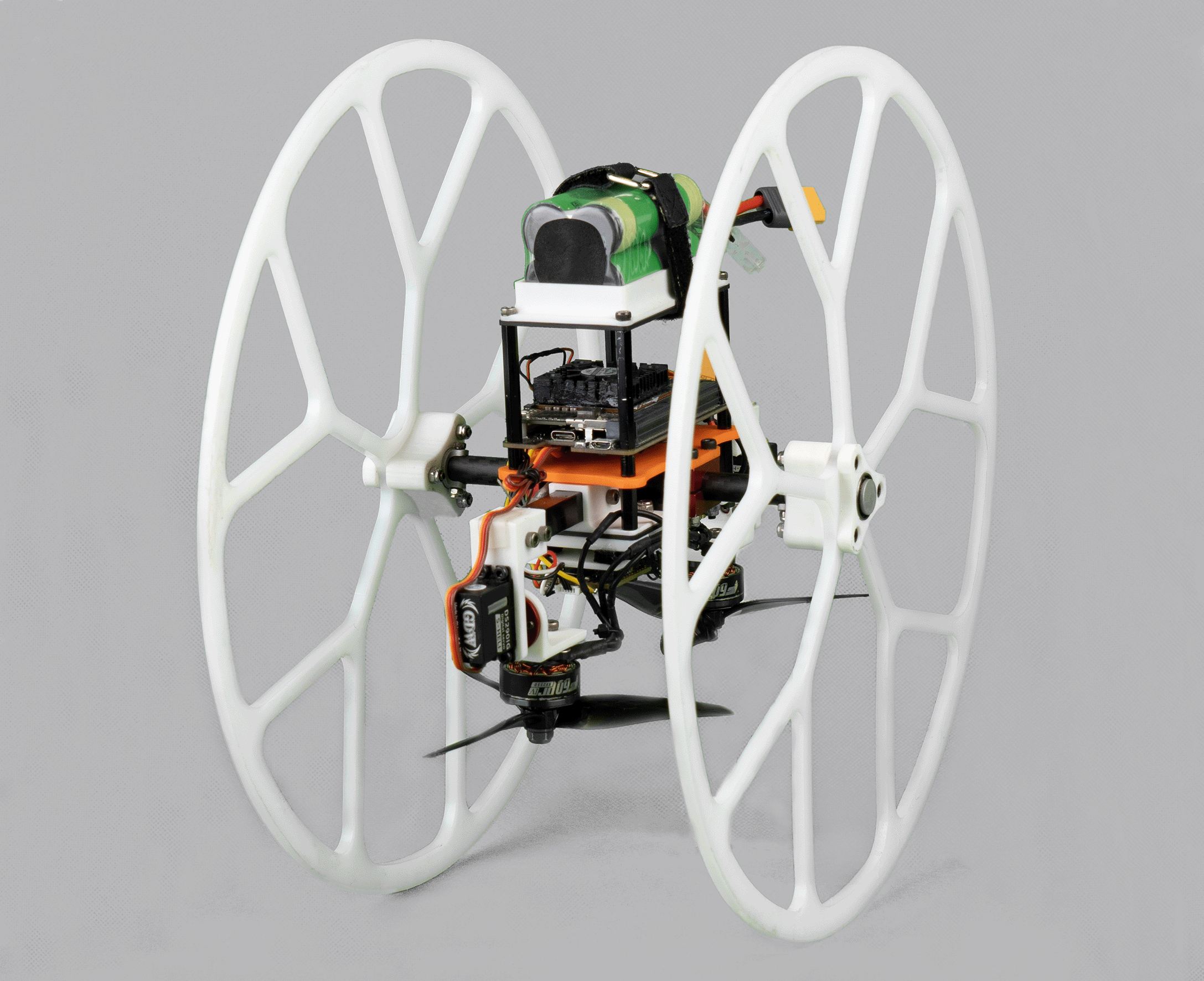}}
		\captionsetup{font={small}}
		\caption{ \label{fig:proposed_TABV}Prototype of the proposed bi-modal bi-copter robot.}
		\vspace{-1.0cm}
	\end{figure}

	Based on the above analysis, we propose a novel bi-copter robot named Skater, as shown in Fig. \ref{fig:proposed_TABV}. The flying mobility of the Skater is achieved through the use of a longitudinally arranged bi-copter, equipping it with excellent traversability. Attaching two passive wheels to both sides of the robot, the Skater employs a unified actuation system in aerial and ground modes. This not only contributes to the structural compactness and lightweight nature of the robot but also provides it with strong steering capacity and the ability to actively generate centripetal force, allowing it to navigate various terrains, whether rough or slippery.
	
	To fully exploit the motion performance, we establish a comprehensive dynamic model for Skater, analyze its differential flatness characteristic, and propose a Nonlinear Model Predictive Control (NMPC) controller for trajectory tracking. Through extensive real-world experiments and benchmark comparisons, we validate the superiority of the robot and the effectiveness of the controller. The main contributions of this paper are as follows:
	
	1) A novel aerial-ground robot with outstanding terrain adaptability and traversability.
	
	2) A comprehensive dynamic model and differential flatness characteristic for the proposed configuration that benefit motion planning and tracking control.
	
	3) A unified NMPC control framework that achieves accurate tracking of aerial and ground trajectories and seamless modal switching.
	
	4) A series of real-world experiments and benchmark comparisons that demonstrate the outstanding performance of the robot and its controller.
	
	\section{Related work} 
	\label{sec:related_works}
	\subsection{Configuration Design}
	\label{sec:related_design}	
	According to the selection of driving mechanisms, aerial-ground robots can be primarily categorized into several types: active-wheeled, deformable, and passive-wheeled.
	
	\cite{kalantari2020drivocopter,tan2021multimodal,cao2023doublebee} install several drive wheels on the multicopters, utilizing different motor drives for aerial and ground movements. This design provides robots with impressive mobility and energy efficiency in ground locomotion mode. However, the added load from the extra actuation system substantially reduces the flight time of robots.  \cite{morton2017small,mintchev2018multi,david2021design,jia2023quadrolltor,sihite2023multi} incorporate a deformable structure to achieve modal transition and multimodal locomotion. Nevertheless, deformable structures typically demand supplementary actuators and intricate mechanisms, leading to an augmented overall weight and heightened structural vulnerability. Additionally, the deformation inevitably introduces an interruption in mode switching. Through the installation of cylindrical cages, spherical cages, or passive wheels on multicopters, \cite{kalantari2013hytaq,zhang2023model,yang2022sytab,dudley2015micro,atay2021spherical,qin2020hybrid,pan2023skywalker} leverage a unified actuation system to achieve multimodal locomotion with a concise and lightweight structure, leading to lower energy consumption in the air. However, their energy efficiency in ground locomotion mode is comparatively lower than that of active-wheeled robots.
	
	\subsection{Control}
	\label{sec:related_control}	
	
	For aerial-ground robots, motion control is essential for achieving autonomy and maximizing the effectiveness of structural design. While the aerial motion control for bi-modal robots with passive wheels is similar to traditional multicopters, ground locomotion requires a distinct control approach due to the interaction with the ground. Currently, research on these robots is predominantly focused on mechanical design, with motion control receiving only preliminary attention. Operations are generally limited to manual control \cite{kalantari2013hytaq,dudley2015micro} or low-speed trajectory tracking\cite{yang2022sytab,jia2023quadrolltor,atay2021spherical}. The challenge of high-speed tracking is not only due to structural constraints but also often stems from oversimplified dynamic models in controller design, frequently neglecting the effects of ground reaction forces, which leads to less than optimal control outcomes.
	
	\cite{zhang2023model} utilizes Incremental Nonlinear Dynamic Inversion (INDI) to estimate ground reaction forces, achieving ground trajectory tracking with a maximum speed of 2 $\mathrm{m/s}$ and a maximum acceleration of 1.8 $\mathrm{m/s^{2}}$. In \cite{pan2023skywalker}, ground support force is represented as an output of differential flatness and defined through piecewise functions. It realizes impressive ground trajectory tracking performance at a maximum speed of 4.5 $\mathrm{m/s}$ and a maximum acceleration of 4.3 $\mathrm{m/s^{2}}$. In this work, we analyze the differential flatness of the robot in ground mode and represent the ground reaction forces using flat outputs and their derivatives. We achieve ground trajectory tracking with a maximum speed of 2.9 $\mathrm{m/s}$ and a maximum acceleration of 3.0 $\mathrm{m/s^{2}}$.
	
	\section{Configuration Design and implementation}
	\label{sec:configuration_design_imple}

	\subsection{Traversability Analysis}
	\label{sec:traver}	 
	
	As previously discussed, the integration of both aerial and ground modes would enlarge the robot's size. To counteract this, the design process should prioritize minimizing the robot's size. In practice, the horizontal width along the vehicle's direction of motion is particularly crucial as it affects the robot's ability to navigate in confined or cluttered space. This section will evaluate the traversability of various multicopters under the same flight efficiency. In this paper, vectors and matrices are indicated in bold, while others are scalars.
		
	\begin{figure}[t]
		\centering  
		{\includegraphics[width=1.0\columnwidth]{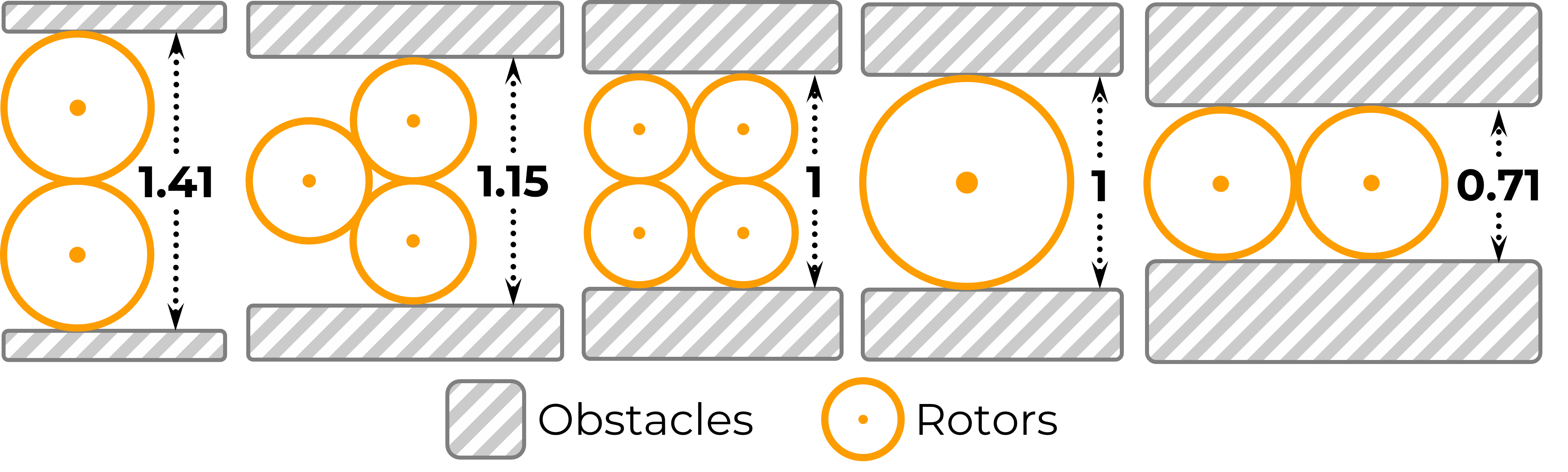}}
		\captionsetup{font={small}}
		\caption{ \label{fig:traver}Theoretical minimum traversing width of common multicopter configurations, which is typically designed to utilize the minimum number of actuators necessary for a specified number of rotors. The width of single-rotor aircrafts is used as the baseline.}
		\vspace{-0.7cm}
	\end{figure}
	Based on the momentum theory \cite{leishman2006principles}, it is known that the ideal power $P$ required to generate a specific thrust $T$ by a rotor is
	\begin{equation}
		\label{eqn_1}
		P(T)=\sqrt{\frac{T^3}{2S\rho } } ,
	\end{equation}
	where $S$ represents the swept area of the rotor, and $\rho$ is the air density. This theory has been experimentally validated in \cite{qin2020gemini}. For rotorcrafts, we typically use hover efficiency $\eta _{h}$ to assess their flight efficiency. In the hover state, the total thrust generated by all rotors equals to its weight. $\eta _{h}$ is defined as
	\begin{equation}
		\label{eqn_2}
		\eta _{h} =\frac{m}{k\times P(mg/k)} = \frac{\sqrt{2kS\rho } }{g\sqrt{mg} } ,
	\end{equation}
	where $m$ is the total mass of the rotorcraft, $k$ is the number of rotors, $g$ is the gravitational acceleration. The radius of a rotor $R$ is
	\begin{equation}
		\label{eqn_3}
		R  = \frac{\eta _{h}g\sqrt{mg}}{\sqrt{2\pi k\rho } } .
	\end{equation}
	
	Equation \eqref{eqn_3} reveals that, when hover efficiency and weight are fixed, the rotor radius is determined by the number of rotors. The theoretical minimum traversing width of multicopters is related to the number and layout of rotors, as shown in Fig. \ref{fig:traver}. It illustrates that among common multicopters, bi-copters moving along the longitudinal direction have the smallest ideal traversing width.
	
	Based on the aforementioned analysis, our proposed robot adopts the longitudinally arranged bi-copter as the flying mechanism. Besides, we adopt a bi-copter configuration with the center of mass above the servo motors. This makes the aerial-ground robot a minimum-phase system in aerial mode \cite{he2022design}, resulting in better attitude control performance compared with conventional bi-copter configuration with the center of mass below the servo motors.
	
	\subsection{Steering Capability Comparison} 
	For passive-wheeled robots, overcoming substantial ground friction during turns poses a challenge for yaw angle control. For a typical quadrotor, the torque on the pitch and roll axes comes from the differential thrust generated by rotors, while the yaw axis torque arises from differential rotor torques, typically an order of magnitude smaller. This characteristic makes yaw angle control more challenging for those passive-wheeled robots using quadrotors as the actuation system. In contrast, for bi-copters, all three-axis torques are generated by the rotor thrust. Consequently, it is intuitive that robots using the bi-copter as the actuation system exhibit superior steering capability compared to those using the quadrotor. In this subsection, we will conduct a detailed analysis to quantify the enhanced steering capability of bi-copters.
	
	To comparably evaluate steering capability of the quadrotor-based vehicle and bicopter-based vehicle under roughly same ground power consumption, we set the total thrust $T_f$ generated by both rotors to the same value less than its weight:
	\begin{equation}
	\label{eqn_4}
	T_{1}^q + T_{2}^q + T_{3}^q + T_{4}^q = T_{f},
	\end{equation}
	\begin{equation}
		\label{eqn_5}
		T_{1}^b + T_{2}^b = T_{f},
	\end{equation}
	where $T^q = [T_{1}^q, T_{2}^q, T_{3}^q, T_{4}^q]^{T} $ represents the thrust generated by the four rotors of the quadrotor, $T^{b} = [T_{1}^b, T_{2}^b]^{T} $ represents the thrust generated by the two rotors of the bi-copter. To simplify calculations, we specifically examine the case where the vehicle rotates in place, ensuring that both pitch and roll axis torques remain zero.
	
	After calculation, the maximum yaw axis torque of the quadrotor-based vehicle is:
	\begin{equation}
		\label{eqn_6}
		\tau _{zmax} ^{q} = \frac{c_q}{c_t}T_f ,
	\end{equation}
	where $c_q$ and $c_t$ are the rotor torque and thrust coefficient.
	When the rotors reach their maximum tilting angle, the maximum yaw axis torque of the bicopter-based vehicle is:
	\begin{equation}
		\label{eqn_7}
		\tau _{zmax} ^{b} = l\cdot T_f ,
	\end{equation}
	where $l$ is the horizontal distance from the rotor center to the center of mass. 
	
	For micro aerial vehicles, $c_q$ is usually in the range of 1$\mathrm{\%}$ to 2$\mathrm{\%}$ of $c_t$, and $l$ is typically on the order of decimeters. Therefore, the steering capability of bicopter-based vehicles is usually more than 5 times stronger than that of quadrotor-based vehicles.
	
	\subsection{Balance Between Energy Efficiency and Agility} 
	In this subsection, we will delve into the reasons behind selecting two passive wheels on both sides for ground locomotion.
	\begin{figure}[t]    
		\centering
		{\includegraphics[width=1.0\columnwidth]{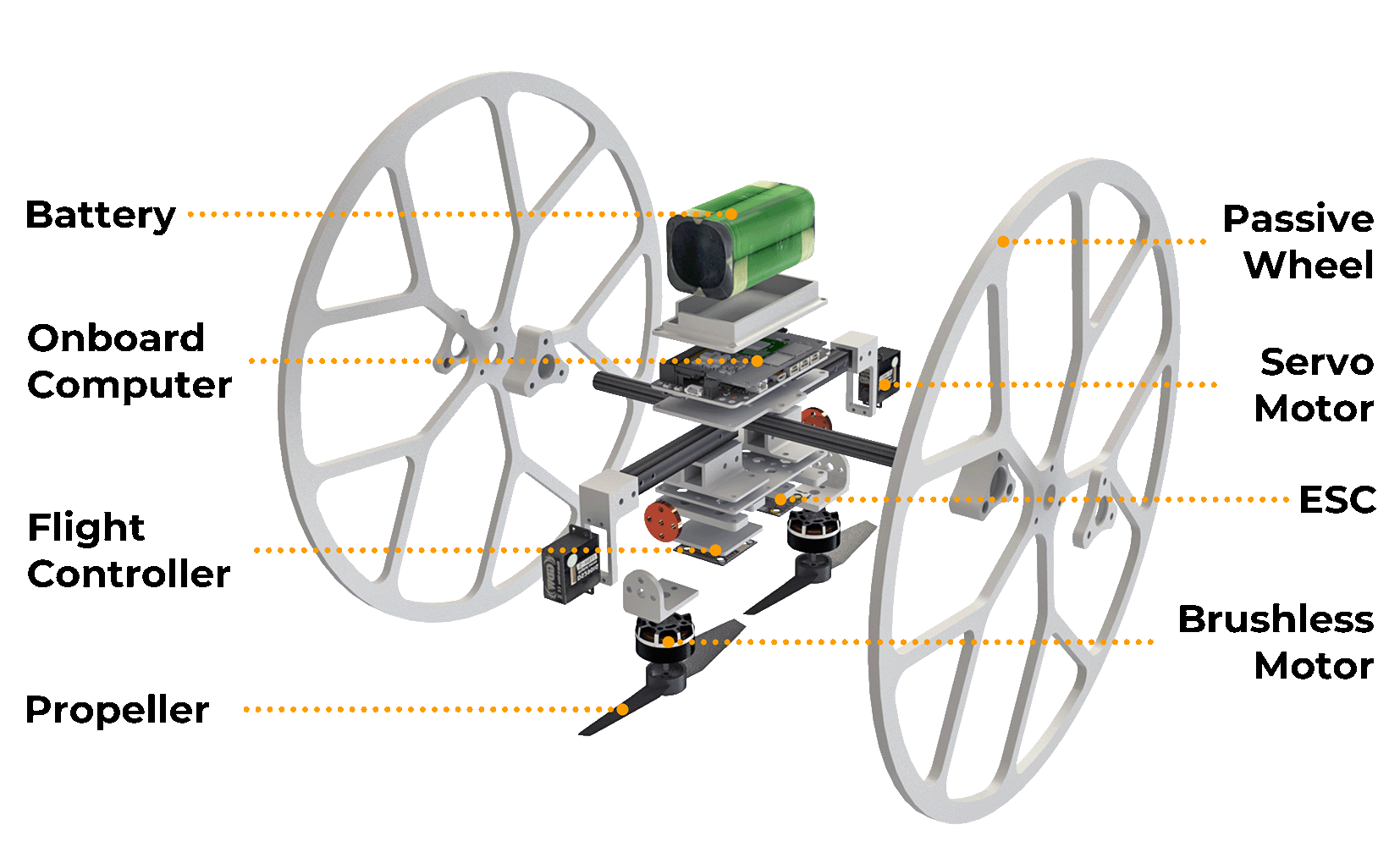}}
		\captionsetup{font={small}}
		\caption{ \label{fig:sw_TABV}Illustration of the hardware details for the proposed system.}
		\vspace{-0.2cm}
	\end{figure}
	\begin{table}[!htb]  
		\renewcommand\arraystretch{1.3}
		\centering
		\caption{Model and weight of key components.}  
		\label{table:1} 
		\begin{tabular*}{\linewidth}{@{}l|l|l@{}} 
			\toprule[2pt]
			\textbf{Component} & \textbf{Model} & \textbf{Weight(g)} \\ \midrule  
			Battery & Sony VTC6 18650 3000mAh 4S & 205  \\ 
			Onboard computer & NVIDIA Jetson Xavier NX & 72  \\  
			Servo motors & GDW DS290IG & 40  \\      
			Brushless motors & T-Motor F60Pro & 66  \\ 
			Flight controller & Holybro Kakute H7 Mini & 8  \\ 
			ESC & Holybro Tekko32 45A & 7  \\ 
			Propellers & GEMFAN 51466 MCK 5-inch & 8  \\ 
			Passive Wheels & Monomer Casting Nylon & 180  \\ \bottomrule[2pt]   
		\end{tabular*}   
		\vspace{-1.0cm}  
	\end{table}
	
	Aerial-ground robots with a single passive wheel can adjust their three-axis attitude on the ground, just as in their aerial mode, allowing for agile omnidirectional movement. However, this agility increases ground energy consumption due to the need for three-axis attitude stabilization. In comparison, those with multiple wheels are more energy-efficient on the ground, as the ground surfaces fix some axes of robots, while their movement can be limited by ground textures.
	
	Our proposed design employs a bi-copter with longitudinal motion as the actuation system, with passive wheels installed on both sides. Although the roll axis attitude is fixed, by rotating two servos in the same direction, it is still possible to generate the required centripetal force for turning. Consequently, this design achieves a balance between high energy efficiency and agility.
	\begin{figure}[ht]  
		\centering
		{\includegraphics[width=0.9\columnwidth]{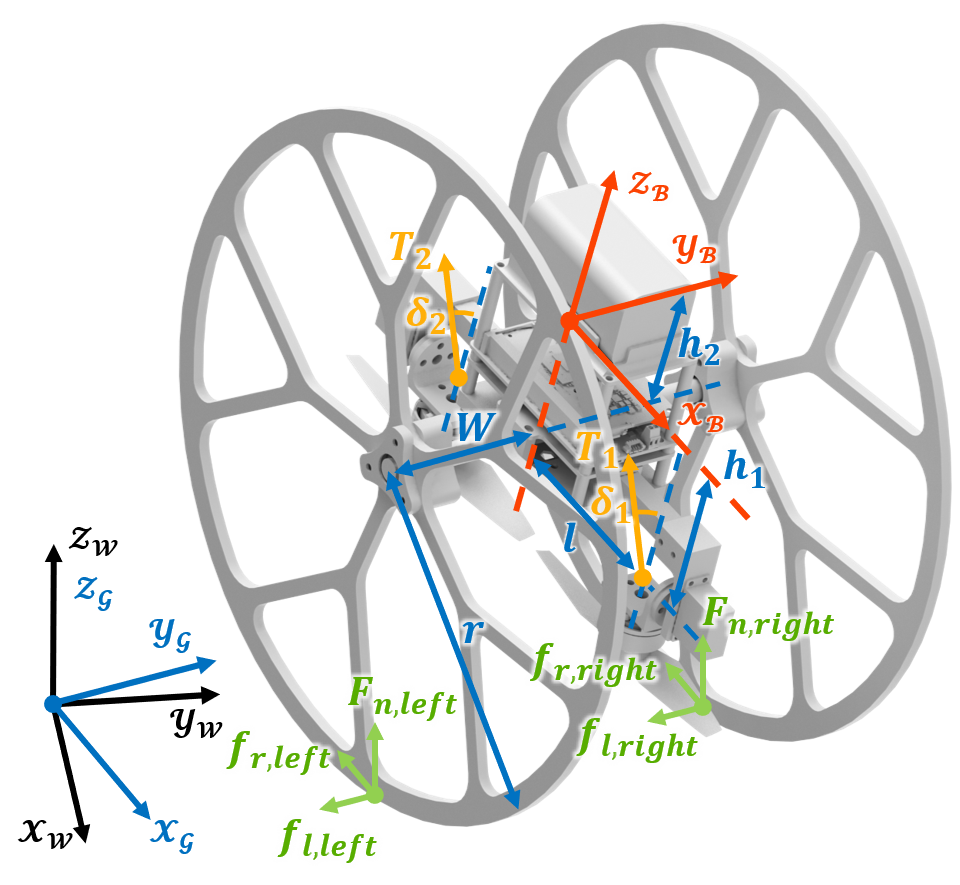}}
		\captionsetup{font={small}}
		\caption{ \label{fig:cord_force}Coordinate definitions and force analysis. $f_r$, $f_l$, and $F_n$ represent the ground rolling friction force, lateral friction force, and normal force on the wheels, with 'left' and 'right' subscripts indicating their application to the left and right wheels.}
		\vspace{-0.2cm}  
	\end{figure}
	
	The illustration of hardware is shown in Fig. \ref{fig:sw_TABV}. The total weight of Skater is 835 $\mathrm{g}$ and the overall size is 18 $\mathrm{cm} \ \times$ 30 $\mathrm{cm} \ \times$ 30 $\mathrm{cm}$. Table. \ref{table:1} lists the model and weight of key components.
	\section{Dynamics and Control}
	\label{sec:dynamic and control}
	\subsection{Unified Dynamics}
	\label{sec:ud}
	Three coordinate frames are used in this paper: the world frame $\bm{\mathcal{F}}_{\mathcal{W} }({\bm{x}_{\mathcal{W} }},\bm{y}_{\mathcal{W} },\bm{z}_{\mathcal{W} })$, the intermediate frame $\bm{\mathcal{F}}_{\mathcal{G} }(\bm{x}_{\mathcal{G} },\bm{y}_{\mathcal{G} },\bm{z}_{\mathcal{G} })$ and the body frame $\bm{\mathcal{F}}_{\mathcal{B} }(\bm{x}_{\mathcal{B} },\bm{y}_{\mathcal{B} },\bm{z}_{\mathcal{B} })$, as shown in Fig. \ref{fig:cord_force}. The world frame  $\bm{\mathcal{F}}_{\mathcal{W} }$ functions as the absolute frame, with its $\bm{z}_{\mathcal{W} }$ oriented vertically upward from the ground. The intermediate frame $\bm{\mathcal{F}}_{\mathcal{G} }$ is derived by rotating $\bm{\mathcal{F}}_{\mathcal{W} }$ about $\bm{z}_{\mathcal{W} }$ until $\bm{x}_{\mathcal{G} }$ aligns with the heading of the robot. The body frame $\bm{\mathcal{F}}_{\mathcal{B} }$ is affixed to the vehicle's Center of Mass (CoM), featuring $\bm{z}_{\mathcal{B} }$ perpendicular to the body and directed upward, while $\bm{x}_{\mathcal{B} }$ consistently follows the vehicle's heading.
	
	The state of the vehicle is represented by $\bm{x} = \{\bm{p},\bm{v},\bm{R},\bm{\omega} \}$, where $\bm{p} = [p_{x},p_{y},p_{z}]^T$ and $\bm{v} = [v_{x},v_{y},v_{z}]^T$ denote the position and velocity of the vehicle's CoM in the world frame, $\bm{R}$ is the rotation matrix representing the orientation in the world frame, and $\bm{\omega} = [\omega_x,\omega_y,\omega_z]^T$ is the angular velocity relative to the body frame. The orientation is also defined by Euler angles of roll, pitch, and yaw ($\phi, \theta, \psi$). The control input is denoted by $\bm{u} = \{T_1,T_2,\delta _1, \delta _2 \}$, where $T_1$ and $T_2$ are thrusts generated by two rotors, and $\delta _1$ and $\delta _2$ are the corresponding servo angles. In the following modeling process, the rotor torque and servo reaction force are neglected due to their relatively small magnitudes compared to other major torques. 
	
	Then, the bi-modal dynamics of system $f(\bm{x},\bm{u})$ is expressed as
	\begin{equation}
		\label{eqn:p_dot}
		\bm{\dot{p}} = \bm{v},
	\end{equation}
	\begin{equation}
		\label{eqn:v_dot}
		\bm{\dot{v}} = \bm{g} + (\bm{R}\bm{T}_{B} + s\bm{R}_{\mathcal{G} }^{\mathcal{W} }\bm{F}_{G})/m,
	\end{equation}
	\begin{equation}
		\label{eqn:R_dot}
		\bm{\dot{R}} = S(\bm{\omega} ) \bm{R},
	\end{equation}
	\begin{equation}
		\label{eqn:w_dot}
		\bm{\dot{\omega}} = \bm{J}^{-1}(-\bm{\omega}  \times \bm{J} \bm{\omega}  + \bm{\tau} _B + s\bm{\tau} _G),
	\end{equation}
	\begin{equation}
		\label{eqn:s(w)}
		S(\bm{\omega}) = \begin{bmatrix}
			0 & -\omega _z &  \omega _y\\
			\omega _z & 0 & -\omega _x\\
			-\omega _y & \omega _x & 0
		\end{bmatrix},
	\end{equation}
	where $m$ is the total mass of the vehicle, $\bm{g} = [0,0,-9.81 \ \mathrm{m/s^2}]^T$,  $\bm{R}_{\mathcal{G} }^{\mathcal{W} }$ represents the rotation from $\bm{\mathcal{F}}_{\mathcal{G} }$ to $\bm{\mathcal{F}}_{\mathcal{W} }$. The switch variable $s = 0$ corresponds to the robot being in the aerial mode, while $s=1$ corresponds to the robot being in the ground mode.  $\bm{T}_{B}$ is the thrusts generated by actuators:
	\begin{equation}
		\label{eqn:t_b}
		\bm{T}_B =
		\begin{bmatrix} {T}_{B,x}
			\\ {T}_{B,y}
			\\ {T}_{B,z}
		\end{bmatrix}
		= \begin{bmatrix}
			0
			\\ -T_1\sin {\delta _1} -T_2\sin {\delta _2}
			\\ T_1\cos {\delta _1} +T_2\cos {\delta _2}
		\end{bmatrix},
	\end{equation}
	$\bm{F}_{G}$ is the ground reaction force in $\bm{\mathcal{F}}_{\mathcal{G} }$:
	\begin{equation}
		\label{eqn:f_g}
		\bm{F_G}= \begin{bmatrix} f_r 
			\\ f_l
			\\ F_n
		\end{bmatrix} 
		= \begin{bmatrix} -\mu (mg - {T}_{B,z}\cos\theta )
			\\ ma_l - {T}_{B,y} 
			\\ mg - {T}_{B,z}\cos \theta 
		\end{bmatrix} ,
	\end{equation}
	where $\mu$ is the coefficient of rolling friction, $a_l = (v_{x}\dot{v_{y}} - v_{y}\dot{v_{x}} )/ \left | \bm{v} \right |$ is the centripetal acceleration. $\bm{J}$ indicates the inertia matrix of the entire robot. $\bm{\tau} _B$ is the torques generated by actuators:
	\begin{equation}
		\label{eqn:tau_b}
		\bm{\tau}_B = \begin{bmatrix}
			\tau_{B,x}
			\\ \tau_{B,y}
			\\ \tau_{B,z}
		\end{bmatrix} = \begin{bmatrix}
			(-T_1\sin{\delta_1} -T_2\sin{\delta_2})h_1
			\\ (-T_1\cos{\delta_1} +T_2\cos{\delta_2})l
			\\ (-T_1\sin{\delta_1} +T_2\sin{\delta_2})l
		\end{bmatrix},
	\end{equation}
	where $h_1$ is the vertical distance from the servo rotation axis to the CoM, $l$ is the arm length. $\bm{\tau} _G$ is the ground reaction torque in $\bm{\mathcal{F}}_{\mathcal{B} }$:
	\begin{equation}
		\label{eqn:tau_g}
		\bm{\tau} _G = {\bm{R}_{\mathcal{G} }^{\mathcal{W} }}^T 
		\begin{bmatrix} f_lr + (F_{n,right}-F_{n,left})W
			\\ (m - 2m_{w})h_{2}g\sin \theta 
			\\ (f_{r,right}-f_{r,left})W
		\end{bmatrix},
	\end{equation}
	where $r$ is the radius of the wheel, $W$ is the horizontal distance from the wheel to the CoM, $m_{w}$ is the mass of the wheel, $h_2$ is the horizontal distance from the wheel axle to the CoM.
	\subsection{Differential Flatness Considering Ground Reaction forces}
	\label{sec:df}
	This section introduces the differential flatness of the vehicle dynamics. In aerial mode, this robot is no different from a conventional bi-copter, whose differential flatness has been well studied in previous work \cite{he2022design}. The flat output is determined by the position $\bm{p}$ and the yaw angle $\psi$ .
	
	In ground mode, we demonstrate that the robot remains differentially flat considering ground reaction forces. Our choice of flat output is
	\begin{equation}
		\label{eqn:flat_ouput}
		\bm{\sigma} =\{\bm{p},T_{B,z}\}.
	\end{equation}
	
	In comparison to the typical choice of flat output for multicopters, we introduce $T_{B,z}$ in (\ref{eqn:t_b}) as a term of the flat output and remove $\psi$, where $T_{B,z}$ is set to be a fixed value less than the total weight of the robot. The detailed flatness transformation is given by
	 \begin{equation}
	 	\label{eqn:flat_trans}
	 	(\bm{x},\bm{u}) = \Psi(\bm{\sigma}).
	 \end{equation}
 	
 	First, the position, velocity, and acceleration are simply the first term of $\bm{\sigma}$, $\dot{\bm{\sigma}}$, and $\ddot{\bm{\sigma}}$, respectively. Then, we show that the attitude, body angular velocity, and control input are also functions of $\bm{\sigma}$ and its derivatives.
 	
 	Due to the heading of the robot is always parallel to the direction of motion, the yaw angle is expressed as
 	\begin{equation}
 		\label{eqn:yaw}
 		\psi = \alpha \cdot arctan2(\dot{p_y},\dot{p_x}),
 	\end{equation}
 	where variable $\alpha =1 $ means forward motion of the vehicle, $\alpha = -1$ means reverse. It is assumed that both wheels of the vehicle maintain contact with the ground, so roll angle $\phi = 0$.
 	
 	Left multiplying (\ref{eqn:v_dot}) by $\bm{x}_{\mathcal{G}}$ gives pitch angle:
 	\begin{equation}
 		\label{eqn:pitch}
 		\theta = \arcsin (\frac{m\bm{\ddot{p}}\cdot \bm{x_{\mathcal{G}}} + \mu mg }{\sqrt{1+\mu^2}T_{B,z} } )-\arctan (\mu).
 	\end{equation}
 	
 	Therefore, the rotation matrix $\bm{R} = \bm{R}_{z}(\psi) \cdot \bm{R}_{y}(\theta) \cdot \bm{R}_{x}(\phi)$ of attitude is determined by Euler angles ($\phi, \theta, \psi$).
 
 	Then substituting $\bm{R}$ and its derivative into (\ref{eqn:R_dot}), we get body angular velocity $\bm{\omega}$:
 	\begin{equation}
 		\label{eqn:w}
 		\bm{\omega } = \begin{bmatrix} \omega _x
 			\\ \omega _y
 			\\ \omega _z
 			
 		\end{bmatrix} = \begin{bmatrix} -\dot{\theta } \sin \psi 
 			\\ \dot{\theta } \cos \psi 
 			\\ \dot{\psi } 
 			
 		\end{bmatrix}.
 	\end{equation}
 
 	The normal forces $F_{n,left},F_{n,right}$ acting on the left and right wheels, in addition to half of $F_n$, also include the vertical forces of the lateral friction torque and $\tau_{B,x}$ on both wheels:
 	\begin{equation}
 		\begin{split}
 			 \label{eqn:f_nl,f_nr}
 			F_{n,left} = \frac{1}{2} F_n - \frac{f_lr}{W} - \frac{\tau _{B,x}\cos \theta }{W} , \\
 			F_{n,right} = \frac{1}{2} F_n + \frac{f_lr}{W} + \frac{\tau _{B,x}\cos \theta }{W} .
 		\end{split}
 	\end{equation}
 
 	Correspondingly, the friction forces acting on the left and right wheels are $f_{r,left} = -\mu F_{n,left}$ and $ f_{r,right} = -\mu F_{n,right} $. By substituting (\ref{eqn:f_nl,f_nr}) and $f_{r,left},f_{r,right}$ into (\ref{eqn:tau_g}), and then replacing (\ref{eqn:tau_b}), (\ref{eqn:tau_g}),  (\ref{eqn:w}) and its derivative into (\ref{eqn:w_dot}), we obtain three equations involving only the flat outputs and their derivatives, as well as four control inputs. The fourth equation relates flat outputs and control inputs is $T_{B,z}$ in (\ref{eqn:t_b}). Finally, the desired control input $\bm{u} = \Psi_u(\bm{\sigma})$ is determined by solving these four equations. We can summarize that the dynamics of the vehicle in ground locomotion mode is differentially flat.
	\subsection{Centripetal Force Generation}
	\label{sec:cfg}
	As mentioned before, the robot can actively generate lateral thrust $T_{B,y}$ to meet the requirement for centripetal force. In this paper, $T_{B,y}$ is determined by the flatness transformation in section \refeq{sec:df} for simplification. After neglecting higher-order terms, rolling friction coefficient, and employing the small-angle approximation, $T_{B,y}$ is given by
	\begin{equation}
		\label{eqn:t_by}
		T_{B,y} \approx \frac{1}{1-h_1/r} ma_l.
	\end{equation}
	
	From equation (\ref{eqn:t_by}), it can be observed that $T_{B,y}$ is roughly proportional to the centripetal force and slightly bigger. Although they are not exactly equal, this is sufficient to enable the robot to move effectively on low-friction surfaces, which is validated in section \ref{sec:experiments}.
	
	\subsection{Nonlinear Model Predictive Control Framework}
	\label{sec:nmpc}
\begin{figure}[t]   
		\centering
		{\includegraphics[width=1.0\columnwidth]{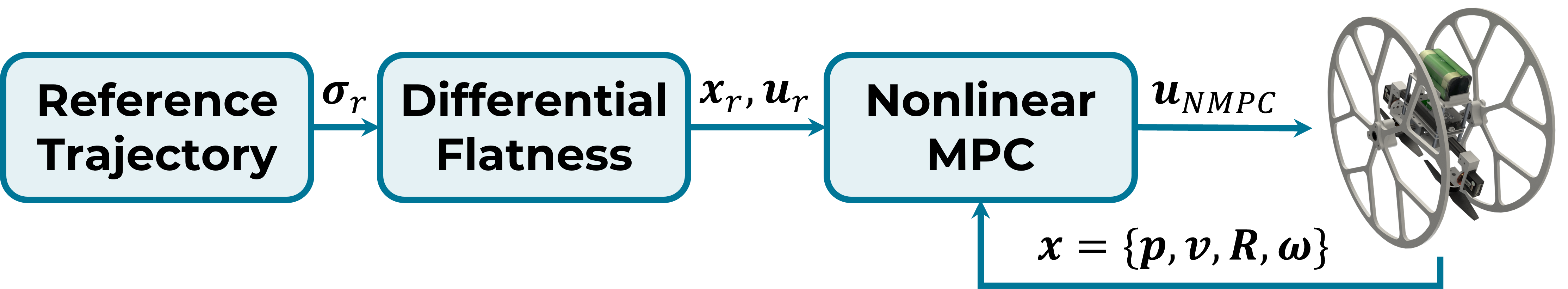}}
		\captionsetup{font={small}}
		\caption{ \label{fig:control_frame}A unified control framework for both modes.}
		\vspace{-0.2cm} 
	\end{figure}

	Thanks to the unified dynamics (\ref{eqn:p_dot}-\ref{eqn:w_dot}) and differential flatness of the vehicle in both modes, we can use a unified controller for both locomotion modes. Accounting for complete dynamics, cascaded PID controller and NMPC controller can achieve similar tracking accuracy on dynamically feasible trajectories \cite{sun2022comparative}. However, the NMPC formulation inherently includes constraints of system dynamics and control inputs while predictively optimizing the control sequence, enabling it to minimize tracking errors on infeasible trajectories while satisfying the constraints. Thus, We use the NMPC controller for trajectory tracking control, and the control framework is shown in Fig. \ref{fig:control_frame}.
	
	\begin{table}[t]  \footnotetext{2}
		\renewcommand\arraystretch{1.3}
		\tabcolsep=0.1cm
		\centering
		\caption{Physical parameters and NMPC gains.}  
		\label{table:2} 
		\resizebox{\linewidth}{!}{
			\begin{tabular*}{\linewidth}{@{}lll|ll@{}} 
				\toprule[2pt]
				\multicolumn{3}{c|}{\textbf{Physical Parmeters}} & \multicolumn{2}{c}{\textbf{NMPC gains}} \\ \midrule  
				$m$, $m_w$   		& $[kg]$       & $0.83$, $0.09$		 & $\bm{Q}_{\bm{p}}$            & $diag(1000,1000,500)$              \\ 
				$\bm{J}$    & $[g\cdot m^2]$   & $diag(4.1,2.8,3.5)$ 	 & $\bm{Q}_{\bm{v}}$      		& $diag(100,100,100)$                  \\ 
				$l$         & $[m]$            & $0.07$ 			 & $\bm{Q}_{\bm{q}}$            & $diag(200,200,200,200)$            \\ 
				$h_1$, $h_2$ & $[m]$            & $0.04$, $0.02$ 			 & $\bm{Q}_{\bm{\omega}}$       & $diag(10,10,10)$                 \\
				$r$, $W$     & $[m]$            & $0.15$, $0.09$ 		 & $\bm{Q}_{\bm{u}}$            & $diag(10,1,1,1)$                 \\
				$c_t$       & $[N\cdot s^2]$    & $1.75e^{-8}$					     &                    			&          \\ \bottomrule[2pt]   
		\end{tabular*}}
		\vspace{-1.5cm} 
	\end{table}
	
	NMPC generates optimal control inputs by solving a nonlinear optimization problem in a receding horizon manner. The optimization problem aims at minimizing a cost function, which measures the errors between predicted states and control inputs and the desired states and inputs in the time horizon $[t,t+K\cdot dt]$. The time horizon is discretized into $K$ equal intervals with a fixed time step $dt$. Then, the NMPC problem is formulated as:
	\begin{equation}
	\begin{split}
		\label{eqn:nmpc}
		\bm{u}_{NMPC}= &\underset{\bm{u}}{min} \sum_{k=0}^{K-1}(\tilde{\bm{x}}(k)^T\bm{Q}\tilde{\bm{x}}(k)+\tilde{\bm{u}}(k)^T\bm{Q}_u\tilde{\bm{u}}(k) ) \\ 
		& \qquad \qquad \qquad \qquad \qquad \quad+\tilde{\bm{x}}(K)^T\bm{Q}\tilde{\bm{x}}(K), \\
		s.t.\ &\bm{x}(k+1) = f(\bm{x}(k),\bm{u}(k)),\ \bm{x}(0) = \bm{x}_{current},\\
		&s\cdot F_{n,left}(k)\ge 0,\ s\cdot F_{n,right}(k)\ge 0, \\
		&\bm{u} \in [\bm{u}_{min}\ \bm{u}_{max}], 
	\end{split}
	\end{equation}
	where $k$ is the current time step, $\tilde{\bm{x}}(k)=\bm{x}(k)-\bm{x}_{r}(k)$ and $\tilde{\bm{u}}(k)=\bm{u}(k)-\bm{u}_{r}(k)$ are the state and input error between estimation and reference, respectively. $\bm{x}_r$ and $\bm{u}_r$ are the reference state and input derived from differential flatness. $\bm{Q} = diag(\bm{Q}_p,\bm{Q}_v,\bm{Q}_q,\bm{Q}_\omega)$,$\bm{Q}_u$ are the weight matrices. $f(\bm{x}(k),\bm{u}(k))$ is the discretized version of (\ref{eqn:p_dot}-\ref{eqn:w_dot}). $\bm{x}_{current}$ is the current state. $F_{n,left}(k)$ and $F_{n,right}(k)$ represent the normal forces acting on the left and right wheels in $k^{th}$ step. $\bm{u}_{min}$ and $\bm{u}_{max}$ are minimum and maximum values of control input. 

	\begin{figure}[t]  
		\centering
		{\includegraphics[width=0.9\columnwidth]{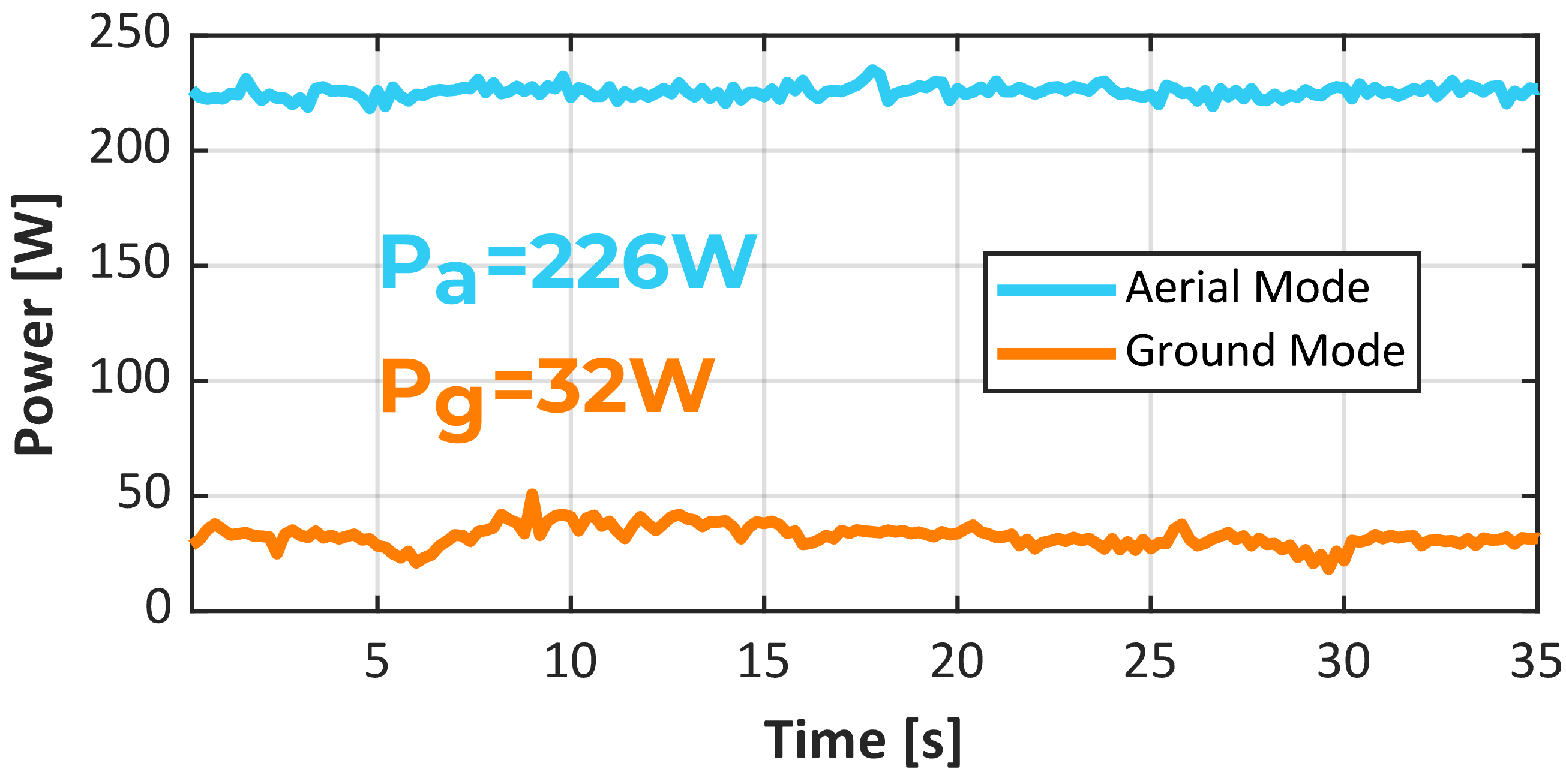}}
		\captionsetup{font={small}}
		\caption{ \label{fig:power_consumption}Power consumption during trajectory tracking in both modes.}
		\vspace{-1.5cm}  
	\end{figure}
	
	ACADO \cite{acado} and qpOASES \cite{qpoases} are used as the solver of the NMPC problem at 200 Hz. $K$ and $dt$ are set to be 20 and 50 $\mathrm{ms}$. The physical parameters and NMPC gains of the robot in real-world experiments are shown in Table. \ref{table:2}.
	\section{Experiments}
	\label{sec:experiments}
	\subsection{Energy Efficiency Validation}
	\label{sec:energy efficiency}
	In this experiment, we have the vehicle track two 8-shaped trajectories in aerial and ground locomotion mode, while recording its power consumption. In all subsequent experiments, the estimation of the robot’s position and attitude are obtained from the motion capture system. The difference between two trajectories is their height, one in the air and the other on the ground. Both trajectories share the same maximum velocity $v_{max}$ and acceleration $a_{max}$ of 1.0 $\mathrm{m/s}$ and 0.6 $\mathrm{m/s^2}$, respectively.

	We measure the standby power $P_s$ of the vehicle, which is 9 $\mathrm{W}$. Subtracting the standby power, the average power of the robot in aerial mode $P_a$ is 226 $\mathrm{W}$, while the average power in ground mode $P_g$ is 32 $\mathrm{W}$, which is shown in Fig. \ref{fig:power_consumption}. Therefore, the energy saving efficiency of the robot in ground mode is $\xi = 1-P_g/P_a=85.8\%$, which means the endurance of the robot in ground locomotion mode is about 7 times longer than in aerial locomotion mode. 
	
	\subsection{Trajectory Tracking Control}
	To demonstrate the effectiveness of the proposed controller, we test the trajectory tracking performance of aerial mode and ground mode respectively, and conduct the tracking experiment of aerial-ground hybrid trajectory. To evaluate the trajectory tracking performance, we choose the root-mean-square-error (RMSE) as the criteria, which is formulated as:
	\begin{equation}
		\label{eqn:rmse}
		RMSE = \sqrt{\frac{\sum_{i=0}^{N} \left \| \bm{p}(i)-\bm{p}_r(i) \right \|^2 }{N} },
	\end{equation}
	where $\bm{p}(i)$ and $\bm{p}_r(i)$ are the $i^{th}$ sampled estimated and desired position of vehicle, respectively.
	
\begin{figure}[t]  
		\centering
		{\includegraphics[width=0.9\columnwidth]{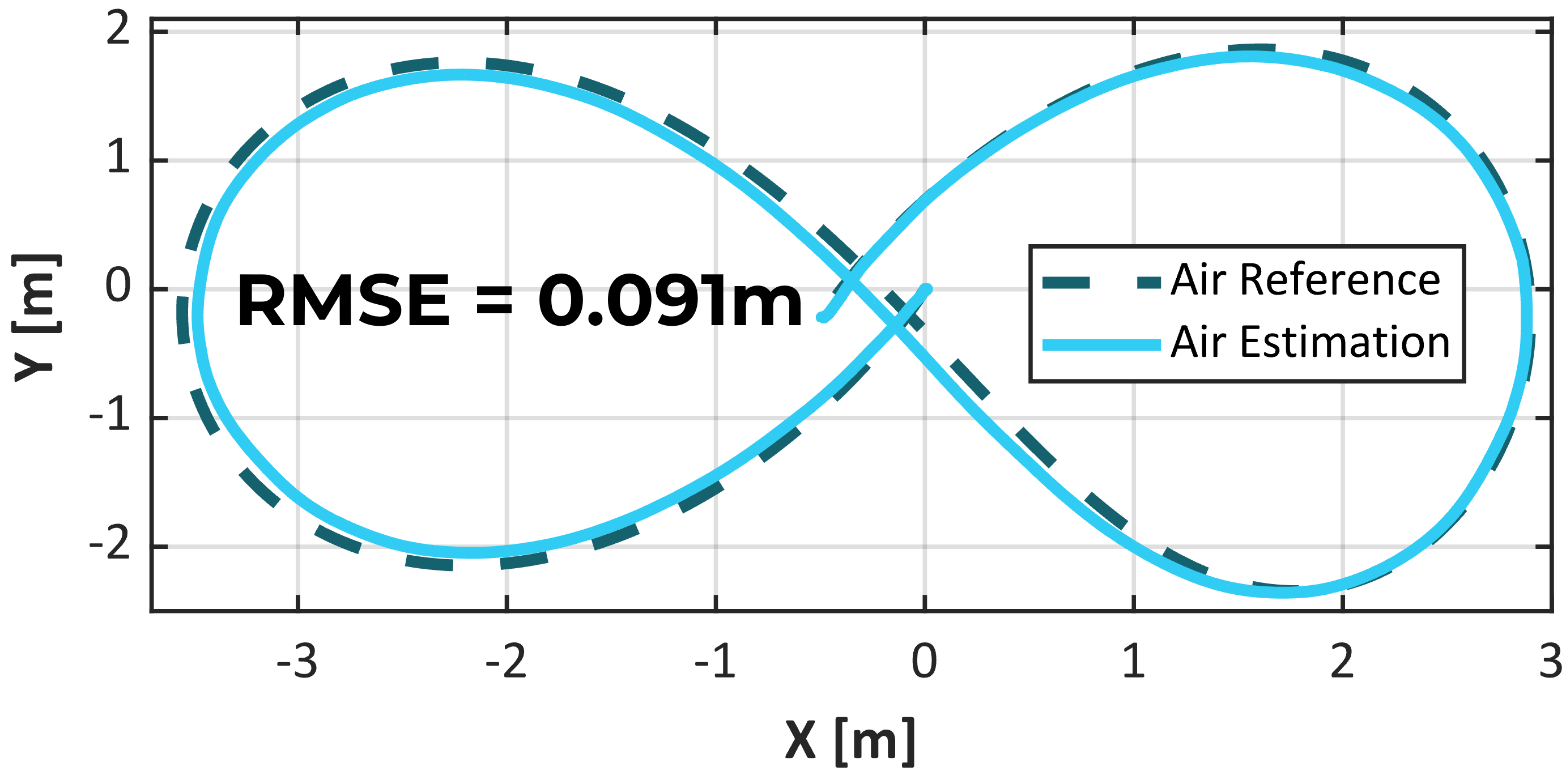}}
		\captionsetup{font={small}}
		\caption{ \label{fig:air_track}Trajectory tracking experiment of aerial locomotion.}
		\vspace{-0.2cm}  
	\end{figure}

	First, a two-dimensional 8-shaped trajectory is executed by the robot in the aerial locomotion mode, with $v_{max}$ at 2.9 $\mathrm{m/s}$ and $a_{max}$ at 3.0 $\mathrm{m/s^2}$. The result is shown in Fig. \ref{fig:air_track}, and the two-dimensional RMSE is 0.091 $\mathrm{m}$.
	
\begin{figure}[t]   
		\centering
		{\includegraphics[width=0.9\columnwidth]{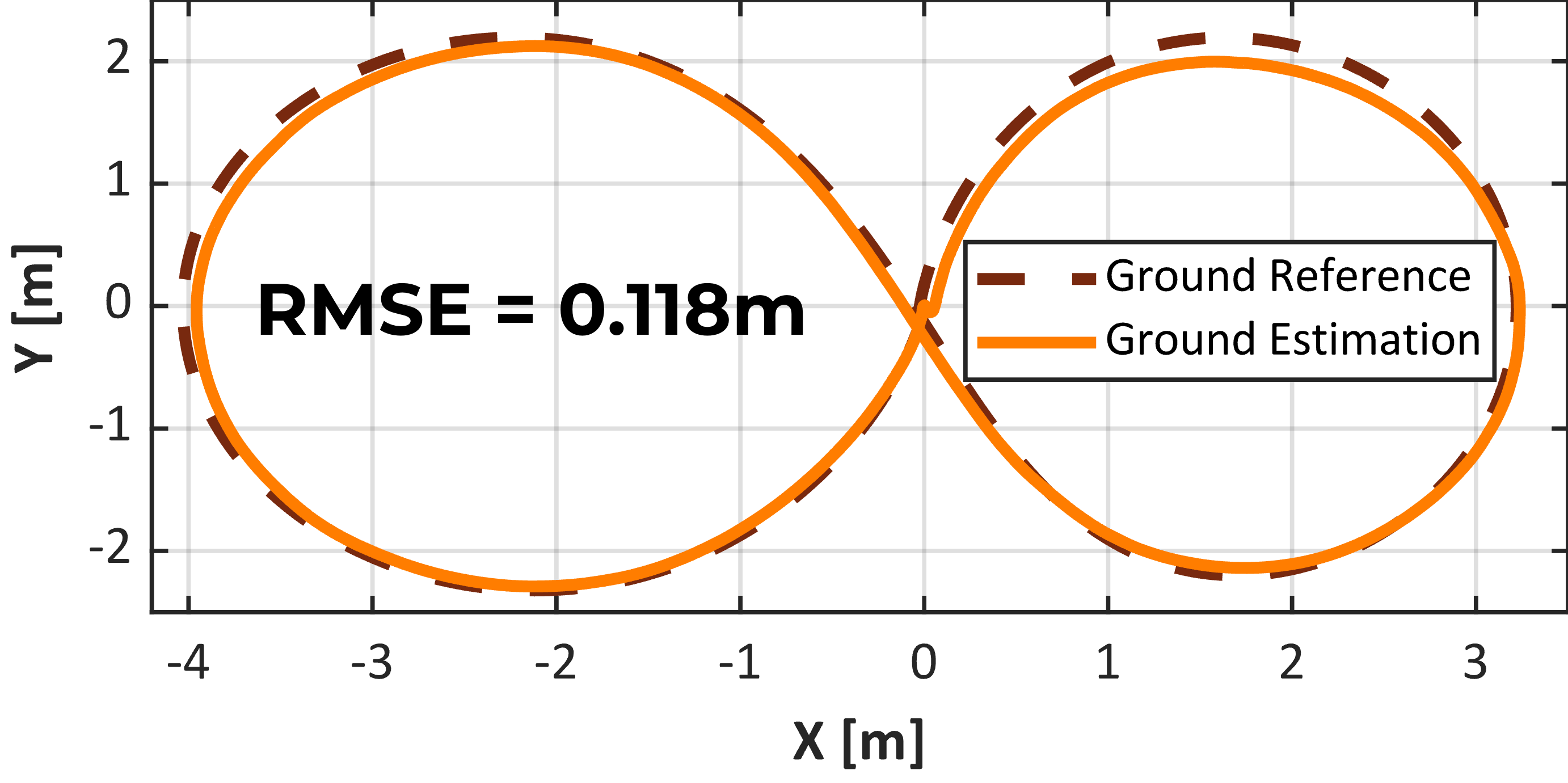}}
		\captionsetup{font={small}}
		\caption{ \label{fig:ground_track}Trajectory tracking experiment of ground locomotion on slippery ground surface.}
		\vspace{-0.2cm} 
	\end{figure}

	Then, to verify the robot's adaptability to diverse ground surfaces, we conduct two separate two-dimensional 8-shaped trajectory tracking tests, one on a slippery surface and the other on a rough surface, as illustrated in Fig. \ref{fig:real_world_experiment}(b)(c). During the test on the slippery surface, $v_{max}$ is set at 2.8 $\mathrm{m/s}$ and $a_{max}$ is 3.0 $\mathrm{m/s^2}$. The trajectory tracking performance is shown in Fig. \ref{fig:ground_track}, with RMSE at 0.118 $\mathrm{m}$. When tracking trajectory on rough surface, $v_{max}$ is 2.9 $\mathrm{m/s}$ and $a_{max}$ is 3.0 $\mathrm{m/s^2}$. The corresponding RMSE for this test also comes at 0.095 $\mathrm{m}$, demonstrating the robot's capability to maintain precise tracking under varying surface conditions.
	
	\begin{figure}[!htb]  
		\centering
		{\includegraphics[width=0.9\columnwidth]{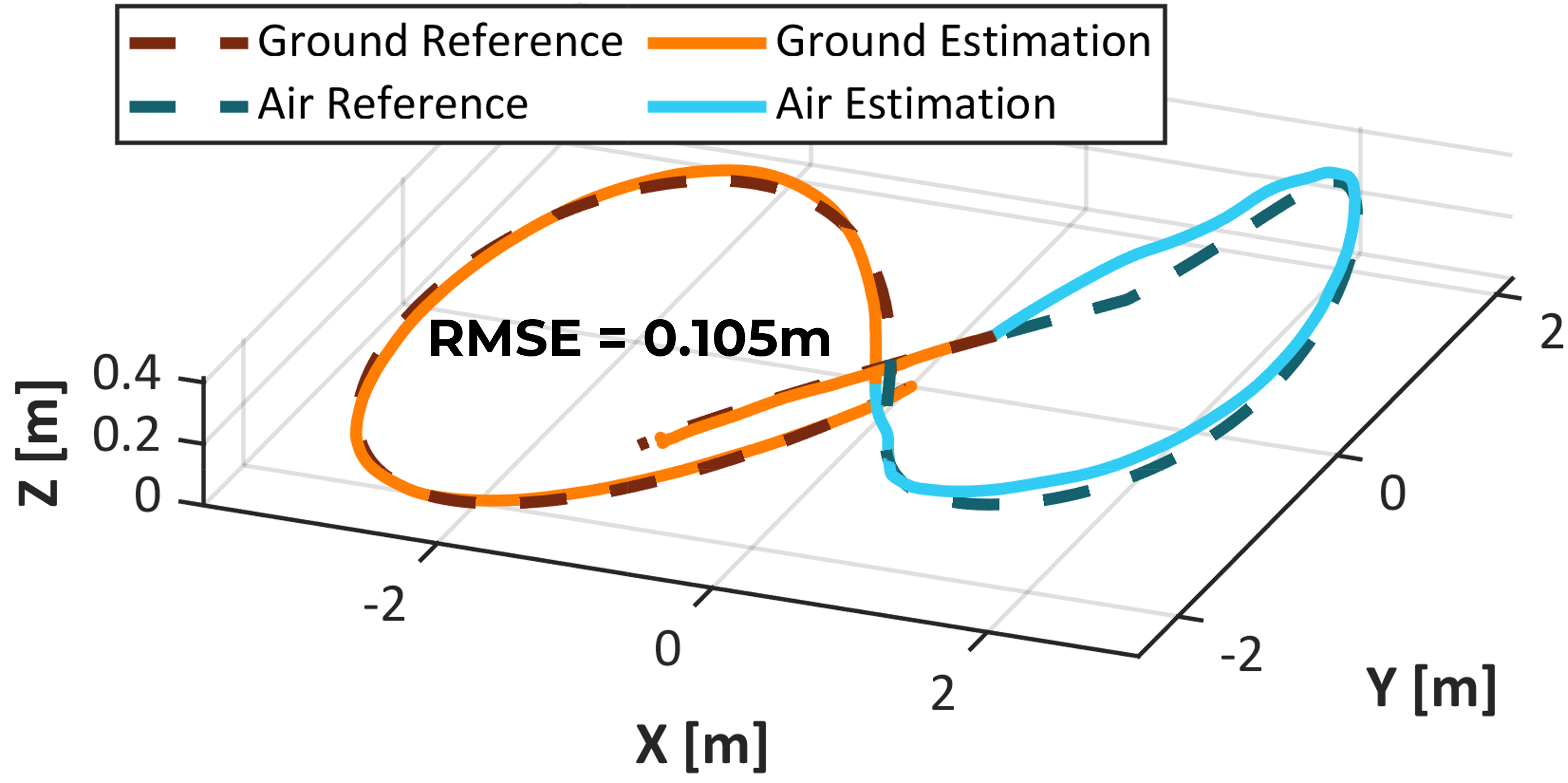}}
		\captionsetup{font={small}}
		\caption{ \label{fig:hybrid_track}Trajectory tracking experiment of hybrid locomotion.}
		\vspace{-1.3cm} 
	\end{figure}

	To demonstrate the smooth modal transition characteristics of the robot, a three-dimensional trajectory tracking test is carried out, as shown in Fig. \ref{fig:real_world_experiment}(a). The upper bounds of velocity and acceleration reach 2.4 $\mathrm{m/s}$ and 2.2 $\mathrm{m/s^2}$, respectively. The result shown in Fig. $\ref{fig:hybrid_track}$ indicates that the proposed controller enables the robot to seamlessly switch modes while tracking trajectories.
	\label{sec:traj_track}
	\begin{figure}[t]   
		\centering
		{\includegraphics[width=0.9\columnwidth]{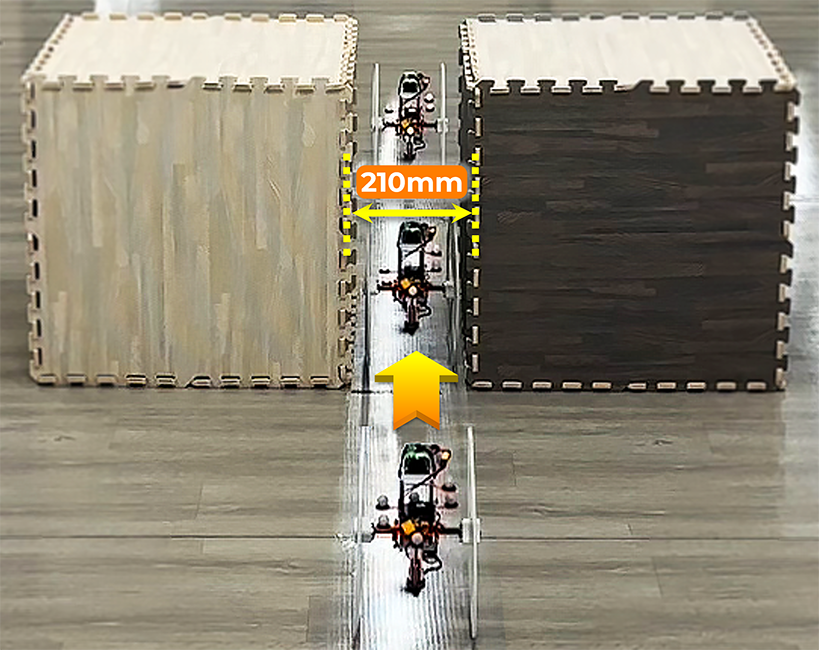}}
		\captionsetup{font={small}}
		\caption{ \label{fig:cross_narrow_gap}Illustration of crossing narrow gap.}
		\vspace{-0.2cm}
	\end{figure}
	\begin{figure}[t]   
		\centering
		{\includegraphics[width=0.9\columnwidth]{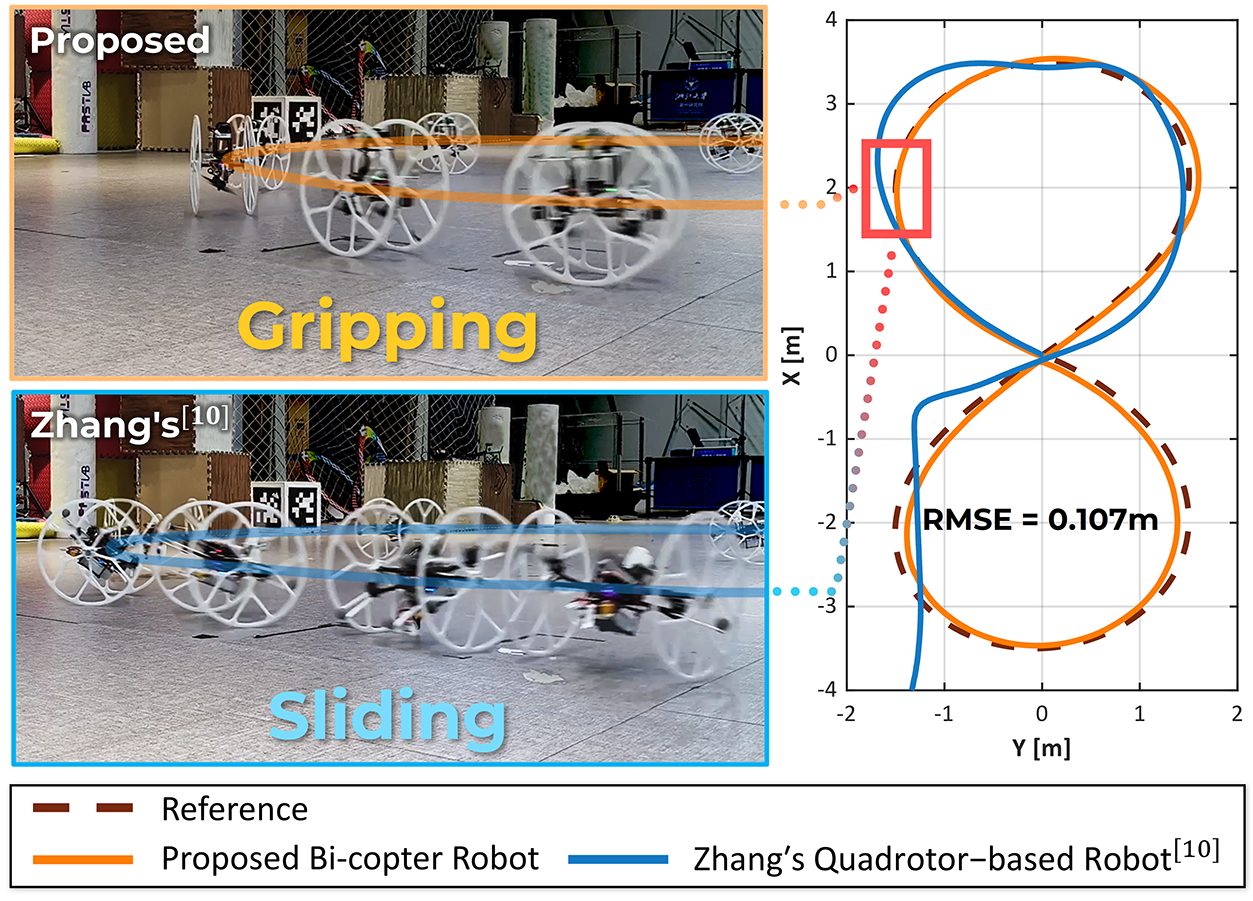}}
		\captionsetup{font={small}}
		\caption{ \label{fig:benchmark}Benchmark comparison of tracking trajectory on the slippery ground surface between the proposed bi-copter robot and Zhang's quadrotor-based robot \cite{zhang2023model}, with maximum speed and maximum acceleration reach 2 $m/s$ and 1.8 $m/s^2$ respectively.}
		\vspace{-0.2cm}
	\end{figure}
	\subsection{Crossing Narrow Gap}
	\label{sec:cross_narrow_gap}
	To validate the terrain traversability of our proposed robot, we conduct an experiment involving navigating through a narrow gap with a width of 210 $\mathrm{mm}$. Fig. \ref{fig:cross_narrow_gap} illustrates the process of the robot traversing through a narrow gap using our proposed NMPC controller, and the maximum velocity reaches 1 $\mathrm{m/s}$.
	
	\subsection{Benchmark Comparisons}
	\label{sec:benckmark}
	To verify the ground textures adaptability of our proposed robot, we compare it to a quadrotor-based aerial-ground robot \cite{zhang2023model} that tracks same trajectories on the same slippery ground surface. When the velocity and acceleration upper bounds reach 1 $\mathrm{m/s}$ and 0.7 $\mathrm{m/s^2}$ respectively, both robots can execute the trajectory completely, while the quadrotor-based robot has the phenomenon of sliding. When the maximum speed and maximum acceleration are increased to 2 $\mathrm{m/s}$ and 1.8 $\mathrm{m/s^2}$ respectively, the centripetal force required for turning is increased. At this point, the slippery ground can not generate enough lateral friction, so the quadrotor-based robot fails. In contrast, our proposed robot can still perform accurate trajectory tracking on this slippery surface with RMSE at 0.107 $\mathrm{m}$, as shown in Fig. \ref{fig:benchmark}.
	
	To emphasize the unique features of our robot and to clearly outline its advantages and disadvantages in comparison to other robots, we conduct a comparative analysis with several representative aerial-ground robots \cite{cao2023doublebee,mintchev2018multi,jia2023quadrolltor,pan2023skywalker,zhang2023model}. The data is mainly derived from corresponding papers. The comparison is summarized in Table. \ref{table:3}, focusing on the following key metrics:
	
	1) Unified actuation system: Robots use same actuators in both aerial and ground locomotion modes.
	
	2) Seamless modal switching: Robots can smoothly switch between two modes without noticeable interruptions.

	3) High-speed autonomy: Robots can perform high-speed ($v_{max} > 1.5 \mathrm{m/s}$) autonomous trajectory tracking.
	
	4) Ground textures adaptability: Robots are able to maintain stable movement on both rough and slippery terrains.
	
	5) Traversing width: The theoretical minimum traversing width of robots, discussed in section \ref{sec:traver}.
	
	6) Energy efficiency: The energy saved by robots in ground mode compared to aerial mode, analyzed in section. \ref{sec:energy efficiency}.
	
	Analyzing Table. \ref{table:3}, we can conclude that while our robot is less energy efficient than active-wheeled robots on the ground, it demonstrates significant advantages in other key performance metrics.

	\section{Conclusion}
	\label{sec:conclusion}
	 In this work, we design and model a novel bi-copter robot, adaptable to air and various ground surfaces. We compare the traversability and steering capabilities of common multicopters and identify the bi-copter, moving along the longitudinal direction with two passive wheels, as the preferred choice for air-ground vehicles to navigate diverse terrains. We then present a dynamic model for the robot, highlighting its differential flatness, which enhances motion planning and control. Subsequently, we introduce a unified NMPC controller for bi-modal trajectory tracking. Real-world experiments and benchmarks confirm the robot's advantages and the controller's effectiveness. Future work should consider extending the model and control of the robot to navigate uneven ground terrains.
	 \begin{table*}[t]  
		\renewcommand\arraystretch{1.3}
		\centering
		\caption{Performance comparison with several aerial-ground robots.} 
		{\includegraphics[width=2.0\columnwidth]{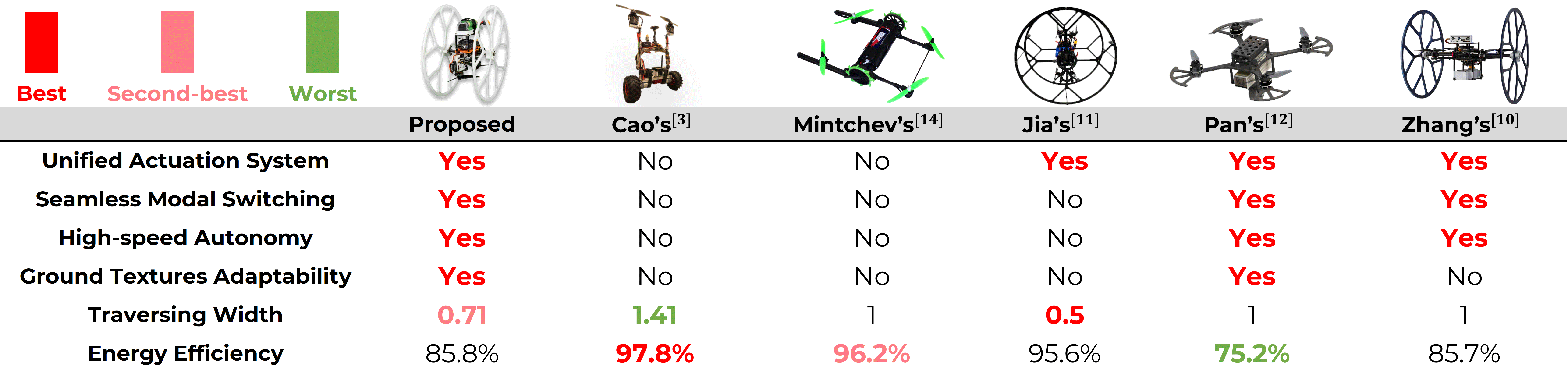}}
		\label{table:3} 
		\vspace{-0.7cm}
	\end{table*}

\bibliography{reference}

\begin{thebibliography}{10}
\providecommand{\url}[1]{#1}
\csname url@rmstyle\endcsname
\providecommand{\newblock}{\relax}
\providecommand{\bibinfo}[2]{#2}
\providecommand\BIBentrySTDinterwordspacing{\spaceskip=0pt\relax}
\providecommand\BIBentryALTinterwordstretchfactor{4}
\providecommand\BIBentryALTinterwordspacing{\spaceskip=\fontdimen2\font plus
\BIBentryALTinterwordstretchfactor\fontdimen3\font minus
  \fontdimen4\font\relax}
\providecommand\BIBforeignlanguage[2]{{%
\expandafter\ifx\csname l@#1\endcsname\relax
\typeout{** WARNING: IEEEtran.bst: No hyphenation pattern has been}%
\typeout{** loaded for the language `#1'. Using the pattern for}%
\typeout{** the default language instead.}%
\else
\language=\csname l@#1\endcsname
\fi
#2}}

\bibitem{kalantari2013hytaq}
A.~Kalantari and M.~Spenko, ``Design and experimental validation of hytaq, a
  hybrid terrestrial and aerial quadrotor,'' in \emph{2013 IEEE International
  Conference on Robotics and Automation}.\hskip 1em plus 0.5em minus
  0.4em\relax IEEE, 2013, pp. 4445--4450.

\bibitem{kalantari2020drivocopter}
A.~Kalantari, T.~Touma, L.~Kim, R.~Jitosho, K.~Strickland, B.~T. Lopez, and
  A.-A. Agha-Mohammadi, ``Drivocopter: A concept hybrid aerial/ground vehicle
  for long-endurance mobility,'' in \emph{2020 IEEE Aerospace
  Conference}.\hskip 1em plus 0.5em minus 0.4em\relax IEEE, 2020, pp. 1--10.

\bibitem{cao2023doublebee}
M.~Cao, X.~Xu, S.~Yuan, K.~Cao, K.~Liu, and L.~Xie, ``Doublebee: A hybrid
  aerial-ground robot with two active wheels,'' \emph{arXiv preprint
  arXiv:2303.05075}, 2023.

\bibitem{sihite2023multi}
E.~Sihite, A.~Kalantari, R.~Nemovi, A.~Ramezani, and M.~Gharib, ``Multi-modal
  mobility morphobot (m4) with appendage repurposing for locomotion plasticity
  enhancement,'' \emph{Nature communications}, vol.~14, no.~1, p. 3323, 2023.

\bibitem{xpeng}
\BIBentryALTinterwordspacing
X.~AEROHT. evtol flying car. [Online]. Available:
  \url{https://intl.aeroht.com/}
\BIBentrySTDinterwordspacing

\bibitem{aeromobil}
\BIBentryALTinterwordspacing
AeroMobil. Am 4.0. [Online]. Available: \url{https://www.aeromobil.com/}
\BIBentrySTDinterwordspacing

\bibitem{pla-v}
\BIBentryALTinterwordspacing
PAL-V. Pla-v liberty. [Online]. Available: \url{https://www.pal-v.com/en}
\BIBentrySTDinterwordspacing

\bibitem{yang2022sytab}
J.~Yang, Y.~Zhu, L.~Zhang, Y.~Dong, and Y.~Ding, ``Sytab: A class of
  smooth-transition hybrid terrestrial/aerial bicopters,'' \emph{IEEE Robotics
  and Automation Letters}, vol.~7, no.~4, pp. 9199--9206, 2022.

\bibitem{tan2021multimodal}
Q.~Tan, X.~Zhang, H.~Liu, S.~Jiao, M.~Zhou, and J.~Li, ``Multimodal dynamics
  analysis and control for amphibious fly-drive vehicle,'' \emph{IEEE/ASME
  Transactions on Mechatronics}, vol.~26, no.~2, pp. 621--632, 2021.

\bibitem{zhang2023model}
R.~Zhang, J.~Lin, Y.~Wu, Y.~Gao, C.~Wang, C.~Xu, Y.~Cao, and F.~Gao,
  ``Model-based planning and control for terrestrial-aerial bimodal vehicles
  with passive wheels,'' in \emph{2023 IEEE/RSJ International Conference on
  Intelligent Robots and Systems (IROS)}.\hskip 1em plus 0.5em minus
  0.4em\relax IEEE, 2023, pp. 1070--1077.

\bibitem{jia2023quadrolltor}
H.~Jia, R.~Ding, K.~Dong, S.~Bai, and P.~Chirarattananon, ``Quadrolltor: A
  reconfigurable quadrotor with controlled rolling and turning,'' \emph{IEEE
  Robotics and Automation Letters}, vol.~8, no.~7, pp. 4052--4059, 2023.

\bibitem{pan2023skywalker}
N.~Pan, J.~Jiang, R.~Zhang, C.~Xu, and F.~Gao, ``Skywalker: A compact and agile
  air-ground omnidirectional vehicle,'' \emph{IEEE Robotics and Automation
  Letters}, vol.~8, no.~5, pp. 2534--2541, 2023.

\bibitem{morton2017small}
S.~Morton and N.~Papanikolopoulos, ``A small hybrid ground-air vehicle
  concept,'' in \emph{2017 IEEE/RSJ International Conference on Intelligent
  Robots and Systems (IROS)}.\hskip 1em plus 0.5em minus 0.4em\relax IEEE,
  2017, pp. 5149--5154.

\bibitem{mintchev2018multi}
S.~Mintchev and D.~Floreano, ``A multi-modal hovering and terrestrial robot
  with adaptive morphology,'' in \emph{Proceedings of the 2nd International
  Symposium on Aerial Robotics}, no. CONF, 2018.

\bibitem{david2021design}
N.~B. David and D.~Zarrouk, ``Design and analysis of fcstar, a hybrid flying
  and climbing sprawl tuned robot,'' \emph{IEEE Robotics and Automation
  Letters}, vol.~6, no.~4, pp. 6188--6195, 2021.

\bibitem{dudley2015micro}
C.~J. Dudley, A.~C. Woods, and K.~K. Leang, ``A micro spherical rolling and
  flying robot,'' in \emph{2015 IEEE/RSJ International Conference on
  Intelligent Robots and Systems (IROS)}.\hskip 1em plus 0.5em minus
  0.4em\relax IEEE, 2015, pp. 5863--5869.

\bibitem{atay2021spherical}
S.~Atay, M.~Bryant, and G.~Buckner, ``The spherical rolling-flying vehicle:
  Dynamic modeling and control system design,'' \emph{Journal of Mechanisms and
  Robotics}, vol.~13, no.~5, p. 050901, 2021.

\bibitem{qin2020hybrid}
Y.~Qin, Y.~Li, X.~Wei, and F.~Zhang, ``Hybrid aerial-ground locomotion with a
  single passive wheel,'' in \emph{2020 IEEE/RSJ International Conference on
  Intelligent Robots and Systems (IROS)}.\hskip 1em plus 0.5em minus
  0.4em\relax IEEE, 2020, pp. 1371--1376.

\bibitem{leishman2006principles}
G.~J. Leishman, \emph{Principles of helicopter aerodynamics with CD
  extra}.\hskip 1em plus 0.5em minus 0.4em\relax Cambridge university press,
  2006.

\bibitem{qin2020gemini}
Y.~Qin, W.~Xu, A.~Lee, and F.~Zhang, ``Gemini: A compact yet efficient
  bi-copter uav for indoor applications,'' \emph{IEEE Robotics and Automation
  Letters}, vol.~5, no.~2, pp. 3213--3220, 2020.

\bibitem{he2022design}
X.~He and Y.~Wang, ``Design and trajectory tracking control of a new bi-copter
  uav,'' \emph{IEEE Robotics and Automation Letters}, vol.~7, no.~4, pp.
  9191--9198, 2022.

\bibitem{sun2022comparative}
S.~Sun, A.~Romero, P.~Foehn, E.~Kaufmann, and D.~Scaramuzza, ``A comparative
  study of nonlinear mpc and differential-flatness-based control for quadrotor
  agile flight,'' \emph{IEEE Transactions on Robotics}, vol.~38, no.~6, pp.
  3357--3373, 2022.

\bibitem{acado}
B.~Houska, H.~J. Ferreau, and M.~Diehl, ``Acado toolkit—an open-source
  framework for automatic control and dynamic optimization,'' \emph{Optimal
  Control Applications and Methods}, vol.~32, no.~3, pp. 298--312, 2011.

\bibitem{qpoases}
H.~J. Ferreau, C.~Kirches, A.~Potschka, H.~G. Bock, and M.~Diehl, ``qpoases: A
  parametric active-set algorithm for quadratic programming,''
  \emph{Mathematical Programming Computation}, vol.~6, pp. 327--363, 2014.

\end{thebibliography}
\end{document}